\documentclass{article}

\PassOptionsToPackage{numbers,compress}{natbib}
\usepackage[preprint]{neurips_2025}

\usepackage{wrapfig}
\usepackage[utf8]{inputenc} 
\usepackage[T1]{fontenc}    
\usepackage{hyperref}       
\usepackage{url}            
\usepackage{booktabs}       
\usepackage{amsfonts}       
\usepackage{nicefrac}       
\usepackage{microtype}      
\usepackage{xcolor}         
\usepackage{times}
\usepackage{latexsym}
\usepackage{multirow}
\usepackage{microtype}
\usepackage{inconsolata}
\usepackage{graphicx}
\usepackage[utf8]{inputenc}
\usepackage{geometry}
\usepackage{array}
\usepackage{booktabs}
\usepackage{multirow}
\usepackage{graphicx}
\usepackage{caption}
\usepackage{makecell}
\usepackage{colortbl}
\usepackage{xcolor}
\usepackage{siunitx}
\usepackage{threeparttable}
\usepackage{resizegather}
\usepackage{amsmath}
\usepackage{amssymb}
\usepackage{bm} 
\usepackage[table]{xcolor}
\usepackage{tcolorbox}
\usepackage{minitoc}

\usepackage{fontawesome}
\usepackage{hyperref}
\usepackage{amssymb}

\usepackage{xspace}

\definecolor{darkred}{RGB}{190, 0, 0}
\definecolor{darkblue}{RGB}{0, 0, 190}

\newcolumntype{L}[1]{>{\raggedright\arraybackslash}p{#1}}

\newcommand{\huggingface}{\raisebox{-1.5pt}{\includegraphics[height=1.05em]{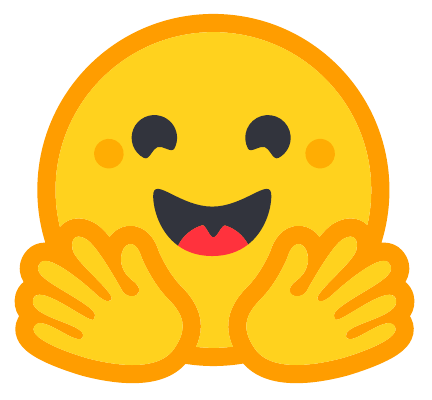}}\xspace}
\newcommand{\github}{\raisebox{-1.5pt}{\includegraphics[height=1.05em]{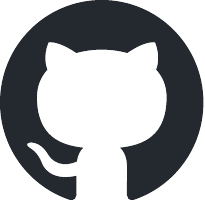}}\xspace}

\title{Forest Before Trees: Latent Superposition for Efficient Visual Reasoning}

\author{%
  Yubo Wang\textsuperscript{$\spadesuit$}\textsuperscript{$\dagger$}\textsuperscript{*},
  Juntian Zhang\textsuperscript{$\clubsuit$}\textsuperscript{*},
  Yichen Wu\textsuperscript{$\diamondsuit$},
  Yankai Lin\textsuperscript{$\clubsuit$},
  Nils Lukas\textsuperscript{$\heartsuit$},
  Yuhan Liu\textsuperscript{$\heartsuit$}\textsuperscript{$\ddagger$}\\
  \textsuperscript{$\heartsuit$}MBZUAI \quad
  \textsuperscript{$\spadesuit$}Fudan University \\
  \textsuperscript{$\clubsuit$}Gaoling School of Artificial Intelligence, Renmin University of China \quad
  \textsuperscript{$\diamondsuit$}Harvard University\\
  \texttt{yubowang25@m.fudan.edu.cn} \quad
  \texttt{zhangjuntian@ruc.edu.cn} \quad
  \texttt{yuhan.liu@mbzuai.ac.ae}
}

\begin{document}

\doparttoc 
\faketableofcontents 

\maketitle

\begin{center}
\vspace{-2em}
\begin{tabular}{rcl}
\github & \textbf{Code} & \href{https://github.com/ybb6/laser}{\path{Laser}}\\[0.5em]
\huggingface & \textbf{Dataset} & \href{https://huggingface.co/datasets/wybb/Laser-ScanPath}{\path{Laser-ScanPath}}\\[0.5em]
\end{tabular}
\end{center}

\begingroup
\renewcommand{\thefootnote}{\fnsymbol{footnote}}
\setcounter{footnote}{0}

\footnotetext[1]{Equal contribution. The order was decided by a coin flip.}

\footnotetext[2]{Work done during an internship at MBZUAI.}

\footnotetext[3]{Corresponding author.}

\endgroup

\begin{abstract}
  While Chain-of-Thought empowers Large Vision-Language Models with multi-step reasoning, explicit textual rationales suffer from an information bandwidth bottleneck, where continuous visual details are discarded during discrete tokenization. Recent latent reasoning methods attempt to address this challenge, but often fall prey to premature semantic collapse due to rigid autoregressive objectives. In this paper, we propose \textbf{\textit{Laser}}, a novel paradigm that reformulates visual deduction via \textit{Dynamic Windowed Alignment Learning(DWAL)}. Instead of forcing a point-wise prediction, Laser aligns the latent state with a dynamic validity window of future semantics. This mechanism enforces a "Forest-before-Trees" cognitive hierarchy, enabling the model to maintain a probabilistic superposition of global features before narrowing down to local details. Crucially, Laser maintains interpretability via decodable trajectories while stabilizing unconstrained learning via \textit{Self-Refined Superposition}. Extensive experiments on 6 benchmarks demonstrate that Laser achieves state-of-the-art performance among latent reasoning methods, surpassing the strong baseline Monet by 5.03\% on average. Notably, it achieves these gains with extreme efficiency, reducing inference tokens by more than 97\%, while demonstrating robust generalization to out-of-distribution domains. We hope this work encourages a paradigm shift from explicit next-token prediction to latent visual reasoning.
\end{abstract}

\section{Introduction}
\label{sec:intro}

\begin{wrapfigure}{r}{0.5\linewidth}
  \centering
  \vspace{-6pt} 
  \includegraphics[width=0.9\linewidth]{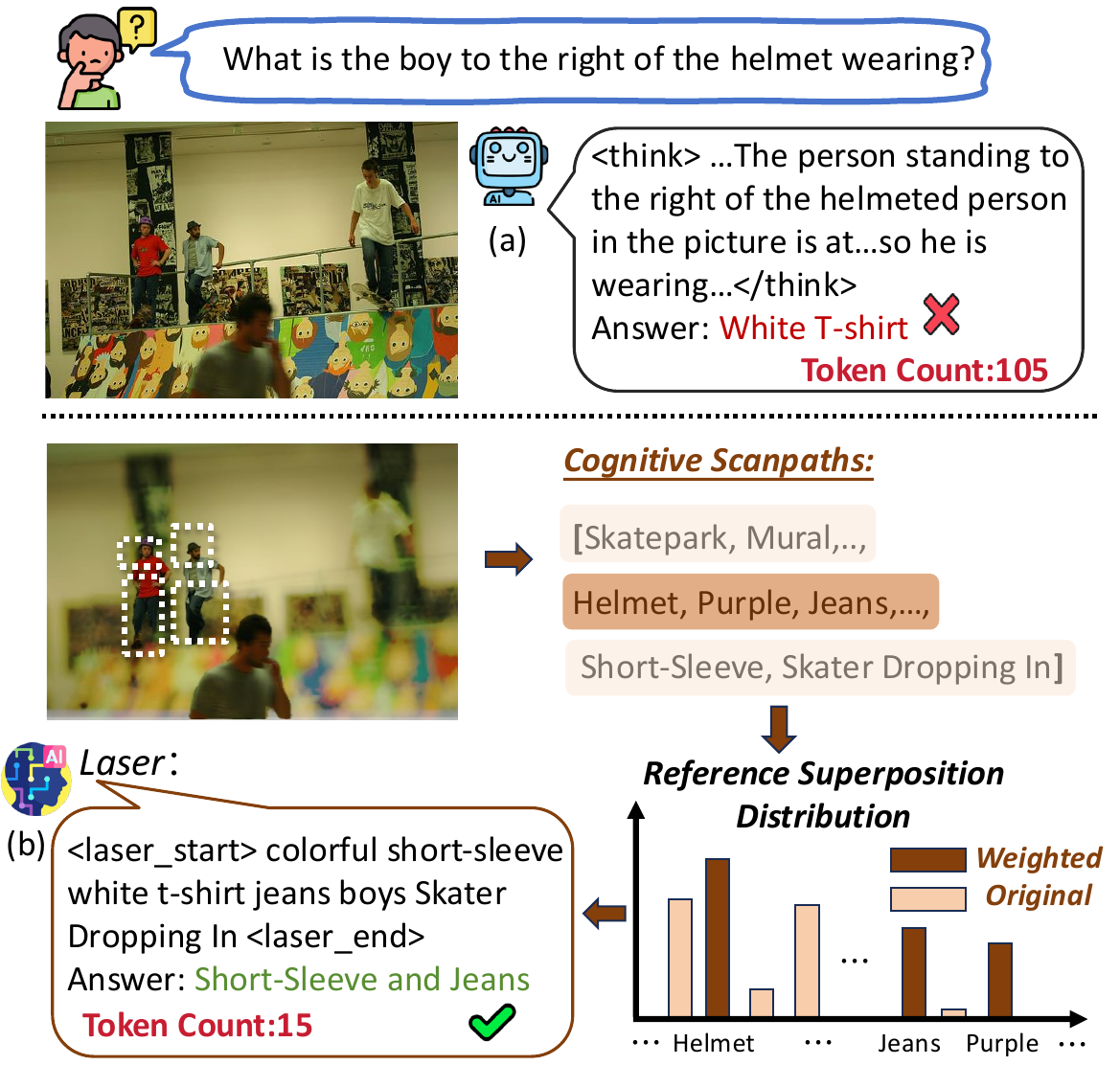}
  \caption{Laser replaces verbose textual rationales (a) with efficient latent superpositions (b).}
  \label{fig:intro}
  \vspace{-6pt}
\end{wrapfigure}

Vision-Language Models (VLMs) have revolutionized visual understanding by integrating Large Language Models (LLMs) with robust visual encoders~\citep{gpt4v, li2024llava}. While adapting Chain-of-Thought (CoT)~\citep{wei2022chain} has enabled multi-step reasoning, as shown in Fig~\ref{fig:intro} (a), explicit textual rationales suffer from an \textit{information bandwidth bottleneck}, where continuous visual details are lost in discrete tokenization~\citep{li2025lvr}. Emerging \textit{Latent Space Reasoning} approaches~\citep{quiet-star,hao2024training} attempt to bypass this by reasoning within high-dimensional hidden states. However, these methods typically retain standard autoregressive objectives, forcing the latent state to strictly minimize prediction error for a specific token at every step. We argue that this strict point-wise mapping is fundamentally misaligned with visual perception. Unlike text generation, visual reasoning is hierarchical, evolving from global semantic apprehension to local feature extraction~\citep{navon1977forest}. Forcing the latent state to prematurely "collapse" into a precise object token before grasping the holistic context induces a \textit{premature semantic collapse}, creating a "tunnel vision" effect that hinders the capture of complex relationships.

To address this gap between sequential objectives and hierarchical perception, we propose \textbf{Laser} (\textbf{La}tent \textbf{S}uperposition for \textbf{E}ffective Visual \textbf{R}easoning). Laser is grounded in the insight that an effective reasoning state should not be a pointer to a single future word, but a container for a \textit{validity domain} of future possibilities. We redefine the optimization goal via \textbf{Dynamic Windowed Alignment Learning (DWAL)}. As shown in Fig~\ref{fig:intro} (b), instead of predicting the immediate next token, Laser trains the latent state to align with a dynamic semantic window encompassing the entire remaining reasoning path. This formulation encourages the latent representation to maintain a \textbf{probabilistic superposition}, simultaneously encoding high-level global semantics while keeping specific details in a potential state. As the reasoning process unfolds, this window naturally shrinks, enforcing a progressive transition from global exploration to local precision, mimicking the general-to-specific nature of human visual processing.

Achieving this "uncollapsed" state without external supervision presents a significant optimization challenge: an unconstrained latent space risks diverging into meaningless high-entropy distributions. First, we introduce \textbf{Self-Refined Superposition}, which leverages the model's own estimation of the future window to construct a stable soft target.Second, to prevent loss of focus, we design an \textbf{Entropy-Regularized Intervention}. This mechanism acts as an implicit curriculum: it dynamically injects rigid ground-truth guidance when the model's uncertainty is high, and reverts to soft superposition when the model demonstrates a grasp of the global context. 
In summary, Laser transforms visual reasoning from a rigid sequential matching task into a flexible, windowed manifold alignment problem. Our extensive experiments on complex visual reasoning benchmarks demonstrate that this approach significantly outperforms both standard CoT and baseline latent reasoning methods. By enabling VLMs to "think" in superpositions before collapsing to answers, Laser bridges the gap between the continuous nature of vision and the discrete nature of language.

Our contributions are summarized as follows:
\begin{itemize}
    \item We propose \textbf{Laser}, a latent reasoning paradigm that reformulates visual deduction via \textbf{Dynamic Windowed Alignment Learning (DWAL)}. This approach prevents \textit{premature semantic collapse} by enforcing a ``forest-before-trees'' cognitive process within the latent space.

    \item We design a supervision framework combining \textbf{Self-Refined Superposition} and \textbf{Entropy-Regularized Intervention}. This establishes an implicit curriculum that stabilizes latent learning without external annotations, dynamically balancing exploration and grounding.

    \item We achieve a superior efficiency-performance balance: Laser obtains state-of-the-art results across 6 benchmarks while reducing inference tokens by over \textbf{97\%}. Furthermore, it demonstrates robust generalization on out-of-distribution tasks, validating the efficacy of the learned visual logic.
\end{itemize}

\section{Related Work}

\subsection{Vision-Language Models}

The evolution of VLMs has rapidly advanced from static cross-modal alignment to dynamic perception.Foundational architectures like Flamingo and BLIP-2 pioneered efficient alignment strategies using Q-Former bottlenecks to bridge frozen vision encoders with Large Language Models (LLMs) \citep{alayrac2022flamingovisuallanguagemodel, li2023blip2bootstrappinglanguageimagepretraining}.Subsequently, 
Open-source models such as LLaVA and MiniGPT-4 demonstrated that simple linear projection layers, coupled with high-quality visual instruction tuning, could achieve strong multimodal following \citep{liu2024visual, zhu2023minigpt4enhancingvisionlanguageunderstanding}. To overcome perceptual limitations in fine-grained and temporal tasks, recent architectures have focused on scaling visual resolution and context length. The latest InternVL3.5, scaled vision encoders to massive parameters using dynamic tiling strategies to handle detailed imagery \citep{chen2024internvl, wang2025internvl35}. Concurrently, Qwen2.5-VL and the advanced Qwen3-VL refined the Naive Dynamic Resolution mechanism, enabling the processing of images at arbitrary aspect ratios and extending capabilities to long-context video understanding \citep{bai2025qwen25vl, bai2025qwen3vltechnicalreport}.
Recent research has focused extensively on enhancing VLMs' reasoning capabilities.
Vision-R1~\citep{huang2025vision} and VL-Rethinker~\citep{wang2025vl} utilize Group Relative Policy Optimization (GRPO) with forced ``caption-reason-answer'' formats and ``rethinking'' tokens
, while VISC focuses on the advancement of multi-image reasoning capabilities~\citep{zhang2025weavingcontextimagesimproving}. In parallel, architectural innovations focus on perceptual self-improvement: ViPER~\citep{zhang2025viperempoweringselfevolutionvisual} establishes a self-evolutionary framework via self-critiquing cycles, and DeepEyes~\citep{zheng2025deepeyes} integrates active perception through dynamic tool invocation. Differently from these approaches, our Laser proposes efficient latent-space reasoning.

\subsection{Latent Space Reasoning}
Explicit Chain-of-Thought enhances model capabilities but suffers from information loss due to discrete tokenization; consequently, LLMs like Quiet-STaR~\citep{zelikman2024quietstar}, Coconut~\citep{hao2024coconut}, and SoftCoT~\citep{xu2025softcot} perform intermediate computations entirely within latent states. In VLMs, the focus shifts to anchoring latent thinking in visual evidence: CoCoVa~\citep{ma2025cocova} and MCOUT~\citep{pham2025mcout} refine representations via latent attention, whereas Mirage~\cite{yang2025mirage}, IVT-LR~\cite{chen2025ivt_lr}, and ILVR~\cite{dong2025ilvr} employ interleaved decoding to stabilize reasoning. While LVR~\cite{li2025lvr} strengthens alignment through autoregressive reconstruction, it risks representation collapse; alternatively, ``visual scratchpads'' like Latent Sketchpad~\cite{zhang2025latentsketchpad} and SkiLa~\cite{tong2025skila} preserve interpretability via reconstructable latent sketches.

Beyond supervised anchoring, other methods directly optimize latent reasoning trajectories (Monet~\cite{wang2025monet}, LaCoT~\cite{sun2025lacot}, DMLR~\cite{liu2025dmlr}). Mull-Tokens~\cite{ray2025mulltokens} offers a modality-agnostic workspace, while Titans~\cite{behrouz2025titans} targets long-range dependencies. Distinct from these, our \textbf{Laser} method introduces a \textit{Dynamic Windowed Alignment mechanism}. By encoding global visual semantics into a compact superposition state, it achieves a superior balance between latent reasoning and efficiency.

\section{Methodology}

In this section, we propose \textbf{Laser}, an efficient visual reasoning method operating in the latent space. Section~\ref{formulation} provides a formal definition of the problem. The key to realizing Laser lies in data acquisition and the design of the training methodology, which are elaborated in Section~\ref{sec:data_construction} and Section~\ref{sec:dynamic_alignment} respectively.

\begin{figure*}[t]
    \centering
    \includegraphics[width=1\linewidth]{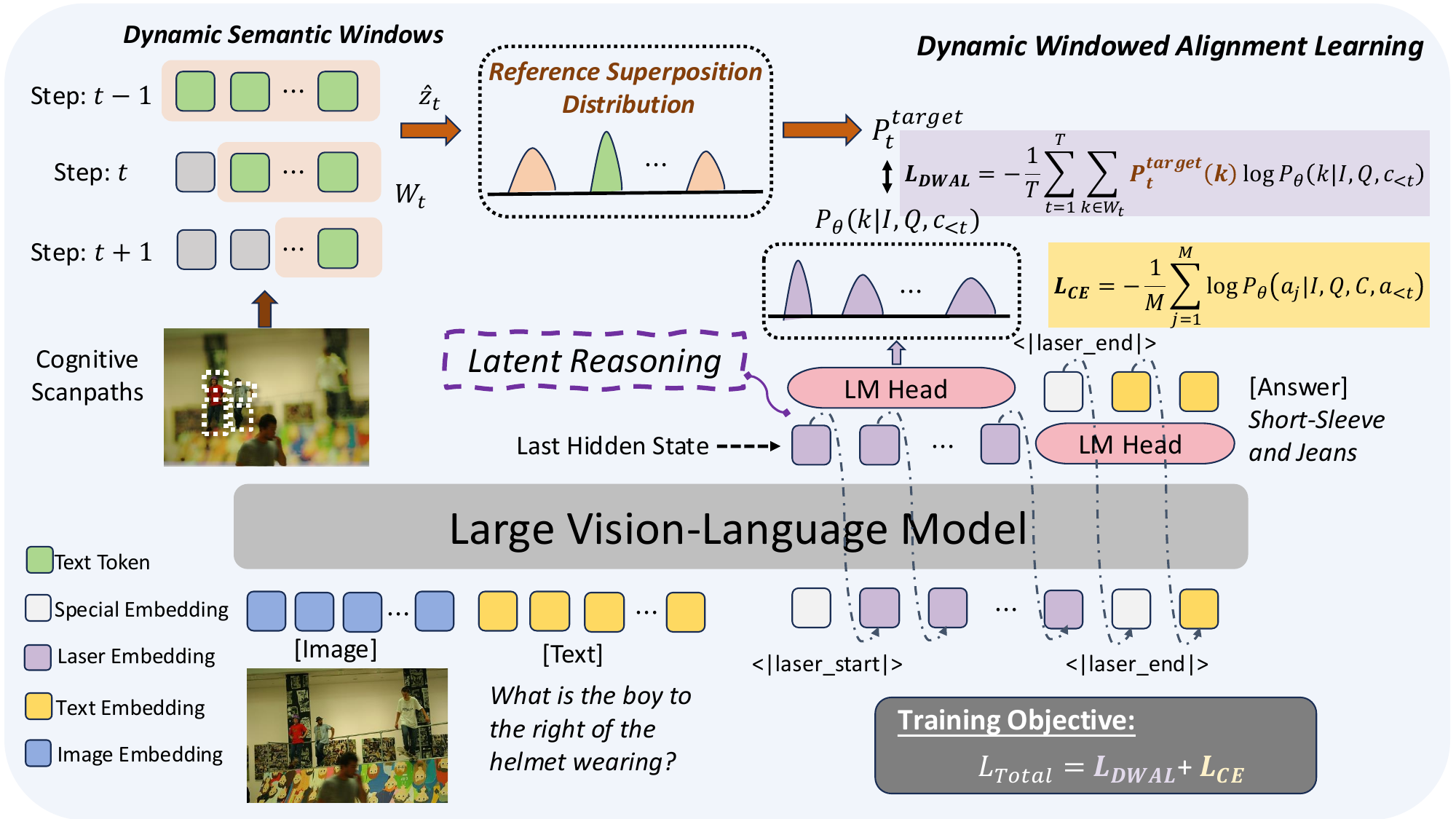}
    \caption{\textbf{Overview of the Laser.} Laser employs \textbf{DWAL}. At each step $t$, a dynamic validity window $\mathcal{W}_t$ is defined over future semantic tokens to construct a \textbf{Reference Superposition Distribution}. The latent state is then optimized to align with this distribution via $\mathcal{L}_{DWAL}$. The final answer is generated explicitly after the reasoning using $\mathcal{L}_{CE}$.}
    \label{fig:method_overview}
\end{figure*}
\subsection{Problem Formulation}
\label{formulation}

We formulate the visual reasoning task as a two-stage conditional generation process: \textit{Latent Visual Reasoning} followed by \textit{Explicit Answer Generation}. Given an input image $I$ and a textual query $Q$, the model $\mathcal{M}_\theta$ aims to synthesize a chain of visual concepts $\mathcal{C} = \{c_1, c_2, \dots, c_T\}$ that acts as an intermediate reasoning path, before producing the final response $\mathcal{A} = \{a_1, a_2, \dots, a_M\}$.

The core of our formulation focuses on the \textit{Latent Reasoning Trajectory}. The model processes the multimodal context to generate a sequence of high-dimensional hidden states $\mathcal{H} = \{h_1, h_2, \dots, h_T\}$, where $h_t = \mathcal{M}_\theta(I, Q, c_{<t})$. 
This latent state is projected onto the vocabulary space via a linear head to obtain the logits $z_t = \mathbf{W}_u h_t$. These logits define the probability distribution over the next token via the Softmax function:
\begin{equation}
    P_\theta(c_{t+1} \mid I, Q, c_{<t}) = \text{Softmax}(z_t).
\end{equation}
In standard autoregressive frameworks, the optimization objective strictly forces each latent state $h_t$ to minimize the negative log-likelihood of the immediate next token $c_{t+1}$. However, this rigid, local constraint compels the latent representation to collapse prematurely into a single semantic point. In this work, we redefine the optimization goal: rather than solely predicting the next token, the latent state $h_t$ is tasked with aligning with a \textit{dynamic validity domain} of future visual semantics. Thus, the problem transforms from minimizing point-wise prediction error to maximizing the \textbf{Windowed Semantic Alignment} between the latent trajectory $\mathcal{H}$ and the reasoning chain $\mathcal{C}$, laying the theoretical foundation for the Dynamic Windowed Alignment Learning (DWAL) detailed in Sec.~\ref{sec:dynamic_alignment}.

\subsection{Synthesizing Cognitive Scanpaths}
\label{sec:data_construction}

For the Laser method, we require a dataset that captures the intermediate visual reasoning process without relying on expensive human annotations. Unlike previous approaches such as Visual CoT~\cite{shao2024}, which anchor reasoning chains with explicit bounding boxes, our aim is to bridge perception and language through \textit{implicit} latent alignment. Consequently, we construct a scalable annotation pipeline that operates under a weakly supervised setting: relying solely on synthesized semantic sequences while deliberately excluding explicit Region-of-Interest (ROI) supervision.

We leverage GPT-4o as a ``Visual Cognitive Engine'' to sequentially synthesize reasoning paths composed of discrete semantic tokens. Crucially, our prompt design is grounded in the \textit{Global Precedence Hypothesis}~\cite{navon1977forest}, which posits that human perception inherently prioritizes holistic structures (``forest'') before processing detailed components (``trees''). Guided by this principle, we enforce a strict \textit{Global-to-Local Scanning Logic}: the synthesized sequences must initiate with a global anchor, progressively narrow focus to relevant objects, and culminate in the critical visual evidence required to answer the query. 
This structure ensures that the data represent valid deductive trajectories rather than static descriptions. After applying a rigorous filtering protocol, we obtain a final dataset named \textbf{\textit{ScanPath}}, consisting of 270k high-quality samples. More details on prompt engineering and filtering statistics are provided in the Appendix~\ref{app:data_details} and~\ref{app:prompt}.

\subsection{Dynamic Windowed Alignment Learning}
\label{sec:dynamic_alignment}

At the core of \textbf{Laser} is the \textbf{Dynamic Windowed Alignment Learning (DWAL)}. Standard autoregressive objectives typically enforce a rigid, point-wise collapse to a single ground-truth token at each step, which we argue is suboptimal for early-stage visual reasoning where global context is paramount. In contrast, DWAL formulates reasoning as a Windowed Probabilistic Alignment problem. By defining a dynamic validity domain of future semantics, this approach allows the latent state to encode a superposition of potential visual concepts, mimicking the general-to-specific nature of human visual processing.

\subsubsection{Dynamic Semantic Windows}
Let $\mathcal{H} = \{h_1, h_2, \dots, h_T\}$ denote the sequence of latent hidden states generated by the model during the reasoning phase, and let $\mathcal{C} = \{c_1, c_2, \dots, c_T\}$ denote the corresponding sequence of text tokens representing visual concepts, which are annotated during data construction.

Previous approaches~\cite{li2025lvr} typically enforce a \textit{Strict Point-wise Mapping}, minimizing the divergence $\mathcal{D}(h_t, c_t)$ between the latent state $h_t$ and the specific ground-truth visual concept $c_t$ at every timestep $t$. To overcome the limitations of this premature semantic collapse, we introduce a \textbf{Dynamic Semantic Window} $W_t$ for each reasoning step $t$. This window defines the ``valid semantic field'' that the current state should encompass:
\begin{equation}
    W_t = \{ c_k \mid t \le k \le T \}.
\end{equation}
The objective is to encourage the latent representation $h_t$ to cover the full spectrum of the valid window $W_t$, rather than peaking solely at the immediate next token $c_{t+1}$. As $t$ increases, the window $W_t$ naturally shrinks ($|W_t| \to 1$), enforcing a progressive transition from global semantic superposition to local precision.

Note that to ensure the reasoning process terminates effectively, the dedicated special token \text{<laser\_end>} is explicitly excluded from the validity window $W_t$ for all steps $t \in [1, T]$. Instead, \text{<laser\_end>} serves as a deterministic target only after the completion of the final reasoning step $T$, signaling the phase transition from implicit reasoning to explicit answer generation.

\subsubsection{Learning via Latent Superposition}
To supervise the model within the dynamic window $W_t$ without relying on external soft labels, we employ a \textbf{Self-Refined Superposition} mechanism. This approach leverages the model's own estimation of the valid semantic manifold to construct a stable soft target. Specifically, we extract the logits corresponding to the tokens in $W_t$ and apply a \textit{stop-gradient} operation to prevent unstable self-reinforcement loops. Let $\hat{z}_t^{(k)} = \text{StopGrad}(z_t^{(k)})$ denote the detached logit for token $k \in W_t$. We define a \textbf{reference superposition distribution} $Q_t$ via a temperature-scaled Softmax:
\begin{equation}
    Q_t(k) = \frac{\exp(\hat{z}_t^{(k)} / \tau)}{\sum_{j \in W_t} \exp(\hat{z}_t^{(j)} / \tau)},
\end{equation}
where $\tau$ is a hyperparameter controlling the sharpness of the distribution. This formulation encourages the hidden state $h_t$ to maintain a probabilistic superposition of future visual concepts.

However, relying solely on soft targets can lead to optimization divergence, where the model converges to a high-entropy uniform distribution lacking semantic focus. To mitigate this, we introduce an \textbf{Entropy-Regularized Intervention}. We first compute the normalized entropy of the reference distribution to gauge the model's uncertainty:
\begin{equation}
H(Q_t) = - \frac{1}{\log |W_t|} \sum_{k \in W_t} Q_t(k) \log Q_t(k).
\end{equation}
We then construct a hybrid target $P^{\text{target}}_t$ that dynamically switches between the soft superposition and a rigid next-token alignment based on this uncertainty:
\begin{equation}
    P^{\text{target}}_t = 
    \begin{cases} 
        \alpha \cdot \mathbf{y}_{\text{hard}} + (1 - \alpha) \cdot Q_t, & \text{if } H(Q_t) > \eta \\
        Q_t, & \text{otherwise}
    \end{cases}
\end{equation}
where $\mathbf{y}_{\text{hard}}$ is the one-hot vector for the immediate next token $c_{t+1}$, $\eta$ is a predefined entropy threshold, and $\alpha \in [0, 1]$ controls the intensity of the hard intervention. This mechanism creates an \textit{implicit curriculum}: it enforces precise grounding when the model exhibits high uncertainty (high entropy), while enabling superposition-based reasoning when the model has grasped the global context.

\subsubsection{Optimization Objective}
The total optimization objective unifies the latent reasoning process and the explicit answer generation. For the reasoning chain, the DWAL Loss minimizes the cross-entropy between the hybrid target and the model's prediction, effectively aligning the latent trajectory with the dynamic semantic windows:
 \begin{equation}
    \mathcal{L}_{\text{DWAL}} = - \frac{1}{T} \sum_{t=1}^{T} \sum_{k \in W_t} P^{\text{target}}_t(k) \log P_\theta(k \mid I, Q, c_{<t}).
\end{equation}
Subsequently, for the answer generation phase, the model produces the final response tokens $\mathcal{A} = \{a_j\}_{j=1}^{M}$ based on the evolved visual understanding. We adopt the standard Cross-Entropy (CE) loss, conditioning on the image $I$, the original query $Q$, and the completed visual chain $\mathcal{C}$:
\begin{equation}
    \mathcal{L}_{\text{CE}} = - \frac{1}{M} \sum_{j=1}^{M} \log P_\theta(a_j \mid I, Q, \mathcal{C}, a_{<j}).
\end{equation}

The final training objective is a weighted summation of these two components:
\begin{equation}
    \mathcal{L}_{\text{Total}} = \mathcal{L}_{\text{DWAL}} +  \mathcal{L}_{\text{CE}}.
\end{equation}
By minimizing $\mathcal{L}_{\text{Total}}$, \textbf{Laser} effectively balances the exploration of global visual semantics during the reasoning phase with the exploitation of precise local semantics for answer generation.

\section{Experiments}

We first demonstrate the superiority of \textbf{Laser} through extensive evaluations on 6 diverse benchmarks against state-of-the-art models. We then conduct further investigations into our method through multi-faceted studies and in-depth analysis.
\subsection{Experimental Setup}
\label{sec:exp_setup}

We instantiate \textbf{Laser} using Qwen2.5-VL-7B-Instruct as the backbone. To preserve the pre-trained visual representations and ensure training efficiency, we freeze the vision tower and the modality merger, exclusively optimizing the LLM parameters. Regarding our specific \textbf{Laser} configuration, we set the temperature $\tau=1.0$ to modulate the softness of the reference distribution, and the entropy threshold $\eta=0.6$ to control the intervention mechanism. Comprehensive hyperparameters are detailed in Appendix~\ref{app:implementation}.

\subsection{Baselines}
\label{sec:baselines}

We evaluate Laser against a comprehensive set of state-of-the-art baselines across three paradigms: (1) \textbf{Zero-shot VLMs} (GPT-4o~\cite{gpt4v}, LLaVA-OneVision~\cite{li2024llava}, InternVL3.5-8B~\cite{wang2025internvl35}, Qwen2.5-VL-7B~\cite{bai2025qwen25vl}), (2) Explicit Visual Interaction methods, including \textbf{tool-augmented reasoning} (DeepEyes~\cite{zheng2025deepeyes}) and \textbf{RL-enhanced VLM reasoning} (Vision-R1~\cite{huang2025vision}, PAPO~\cite{wang2025perception}, VL-Rethinker~\cite{wang2025vl}, and (3) \textbf{Latent VLM Reasoning} approaches (LVR~\cite{li2025lvr}, Monet~\cite{wang2025monet}). Please refer to Appendix~\ref{app:baseline} for more details.

\subsection{Benchmarks}
\label{sec:benchmarks}
We evaluate Laser on six comprehensive benchmarks covering diverse scenarios.For \textit{visual perception}, we use \textbf{BLINK} \cite{blink} to stress image-dependent perception such as depth and spatial cues, and \textbf{MMVP}~\cite{mmvp} to probe CLIP-blind visual patterns. For \textit{visual reasoning}, we adopt \textbf{MMStar}~\cite{mmstar}, which evaluates fine-grained reasoning axes while minimizing shortcut leakage. To probe \textit{high-resolution understanding}, we include \textbf{HRBench}~\cite{hrbench} for ultra-high-resolution (4K only) visual perception. Moreover, we assess \textit{trustworthiness and text-rich understanding} using \textbf{HallusionBench}~\cite{hallusionbench} to diagnose visual illusion and language hallucinations (measured by Question Accuracy, Q-Acc), and \textbf{SEED-Bench-2-Plus}~\cite{seedbench2plus} to evaluate comprehensive understanding of text-intensive visuals such as charts, maps, and web pages.
Each benchmark are detailed in Appendix~\ref{app:benchmark}.
\subsection{Results}
\label{sec:main_results}

\begin{table*}[htbp]
\centering
\setlength{\tabcolsep}{6pt}
\renewcommand{\arraystretch}{1.08}

\definecolor{lightblue}{RGB}{223,234,242}
\definecolor{lightpink}{RGB}{255,230,235}
\definecolor{graytxt}{RGB}{100,100,100}
\definecolor{categorygray}{RGB}{235,235,235}

\resizebox{\textwidth}{!}{%
\begin{tabular}{lccccccc}
\toprule
Model 
& MMVP 
& BLINK 
& SEEDBENCH2PLUS 
& MMStar 
& Hallusion Bench 
& HRBENCH 
& Overall \\
\midrule

\rowcolor{categorygray}
\multicolumn{8}{c}{\textbf{\textit{Zero-Shot VLMs}}} \\ 
GPT-4o~\cite{gpt4v}
& 68.70 & 68.00 & 72.00 & 64.70 & -- & -- & -- \\

Qwen2.5-VL-7B~\cite{bai2025qwen2}
& 65.67 & 53.60 & 65.31 & 59.70 & 56.57 & 68.25 & 61.52 \\

LLaVA-OneVision~\cite{li2024llava}
& 74.00 & 49.34 & 61.22 & 59.13 & 51.10 & 63.00 & 59.63 \\

InternVL3.5-8B~\cite{wang2025internvl3}
& 57.67 & 54.81 & 69.78 & 53.33 & 56.15 & 59.38 & 58.52 \\

\midrule

\rowcolor{categorygray}
\multicolumn{8}{c}{\textbf{\textit{Tool-use \& RL Enhanced Reasoning}}} \\
PAPO~\cite{wang2025perception}
& 68.67 & 52.66 & 54.11 & 45.80 & 57.52 & 68.12 & 57.81 \\

Vision-R1~\cite{huang2025vision}
& 72.67 & 52.71 & 68.95 & 62.67 & 63.83 & 75.12 & 65.99 \\

VL-Rethinker~\cite{wang2025vl}
& 72.67 & 55.55 & 70.27 & 63.20 & 71.08 & 63.50 & 66.05 \\

DeepEyes~\cite{zheng2025deepeyes}
& 70.00 & 51.08 & 69.08 & 58.73 & 62.57 & 69.12 & 63.43 \\

\midrule

\rowcolor{categorygray}
\multicolumn{8}{c}{\textbf{\textit{Latent Reasoning}}} \\
Monet~\cite{wang2025monet}
& \underline{68.00} & 50.71 & \underline{65.88} & \textbf{60.33} & 56.36 & \underline{68.00} & \underline{61.55} \\

LVR~\cite{li2025latent}
& 64.00 & \underline{53.60} & 47.39 & 57.93 & \underline{65.19} & 53.62 & 56.96 \\

\rowcolor{lightblue}
\textbf{Laser (Ours)}
& \textbf{72.00} & \textbf{56.92} & \textbf{70.05} & \underline{60.27} & \textbf{67.72} & \textbf{72.50} & \textbf{66.58} \\

\midrule
\rowcolor{lightpink}
\multicolumn{1}{c}{$\Delta\uparrow$}
& 4.00 & 6.21 & 4.17 & -0.33 & 11.36 & 4.50 & 5.03 \\

\bottomrule
\end{tabular}
}
\caption{Main results comparing \textbf{Laser} with baselines across three paradigms: Zero-Shot VLMs, Tool-use \& RL, and Latent Reasoning. The best results among \textit{latent reasoning} methods are highlighted in \textbf{bold}, and the second best are \underline{underlined}. $\Delta$ denotes the absolute performance gain over the strongest latent baseline, Monet~\cite{wang2025monet}.}
\label{tab:results}
\end{table*}

The comparative results across six benchmarks are presented in Table~\ref{tab:results}. Laser establishes a new state-of-the-art among latent reasoning methods and demonstrates superior capabilities even against computationally intensive explicit reasoning paradigms.

As our primary focus, Laser significantly outperforms existing latent reasoning baselines. Compared to the previous best method, Monet, Laser achieves a remarkable +5.03\% gain in the overall score. Notably, we observe the most substantial improvements on HallusionBench (+11.36\%) and BLINK (+6.21\%). We attribute these gains to the proposed Dynamic Windowed Alignment. By maintaining a semantic superposition rather than collapsing to a rigid token prematurely, Laser effectively mitigates the hallucination issues common in point-wise latent methods and captures fine-grained visual details. In contrast, LVR, which enforces strict next-token reconstruction, lags significantly behind (-9.62\%), highlighting the necessity of our flexible windowed strategy.

Beyond the latent domain, Laser compares favorably against heavyweight paradigms. Despite operating purely in the compact latent space without external tools or iterative reinforcement learning search, Laser surpasses both the leading RL-based method, Vision-R1, and the tool-augmented VL-Rethinker. This suggests that optimizing the internal cognitive trajectory via superposition is a more efficient path to enhanced reasoning than externalizing the process into lengthy textual chains.
Laser consistently outperforms its backbone model, Qwen2.5-VL-7B, across all evaluated benchmarks. Specifically, on the MMVP benchmark, which tests CLIP-blind patterns, Laser improves upon the baseline by +6.33\%. This demonstrates that our method is effectively unlocks latent visual discrimination capabilities that are otherwise dormant in standard supervision settings.

\section{Discussions}
In this section, we address five key research questions through systematic investigation, providing deeper insights into Laser's performance and behavior.

\noindent \textbf{RQ1: How efficient is Laser compared to baselines?}

\begin{table}[htbp]
    \centering
    \resizebox{0.75\linewidth}{!}{
    \begin{tabular}{lcccc}
        \toprule
        \multirow{2}{*}{\textbf{Model}} & \multicolumn{2}{c}{\textbf{Blink} ($N$=1901)} & \multicolumn{2}{c}{\textbf{HrBench} ($N$=800)} \\
        \cmidrule(lr){2-3} \cmidrule(lr){4-5}
         & Avg Tokens & $\Delta$ & Avg Tokens & $\Delta$ \\
        \midrule
        
        \rowcolor{yellow!10}
        Qwen2.5-VL-7B    & 223.5 & --       & 55.9  & --       \\
        \rowcolor{yellow!10}
        VL-Rethinker & 207.0 & \textcolor{darkred}{$\downarrow$-7.4\%}   & 143.8 & \textcolor{darkblue}{$\uparrow$ +157.2\%} \\
        
        \rowcolor{blue!7}
        Monet-7B     & 118.3 & \textcolor{darkred}{$\downarrow$-47.1\%}  & 86.8  & \textcolor{darkblue}{$\uparrow$+55.3\%}  \\
        \rowcolor{blue!7}
        LVR          & 8.0   & \textcolor{darkred}{$\downarrow$-96.4\%}  & 8.0   & \textcolor{darkred}{$\downarrow$-85.7\%}  \\
        \rowcolor{blue!7}
        \textbf{Laser}         & \textbf{6.0}   & \textcolor{darkred}{\textbf{$\downarrow$-97.3}\%}  & \textbf{5.7}   & \textcolor{darkred}{\textbf{$\downarrow$-89.7}\%}  \\
        \bottomrule
    \end{tabular}
    }
    \caption{Efficiency comparison on Blink and HrBench. Our \textbf{Laser} achieves a significant reduction in token usage. The \colorbox{yellow!10}{light yellow} background indicates \textit{Explicit Reasoning} methods, while the \colorbox{blue!7}{light blue} background represents \textit{Latent Reasoning} methods. $N$ denotes the total number of samples, $\Delta$ represents the relative change compared to the Qwen2.5-VL-7B, and \textcolor{darkred}{$\downarrow$} indicates efficiency improvement.}
    \label{tab:token_comparison}
\end{table}

Beyond raw accuracy, the practical deployment of VLMs is often constrained by inference latency and computational cost. We analyze the inference efficiency of \textbf{Laser} compared to both standard baselines and recent latent reasoning methods on the BLINK and HRBench. The results are summarized in Table \ref{tab:token_comparison}.
Existing explicit reasoning approaches, such as VL-Rethinker, typically rely on generating lengthy textual rationales to bridge the gap between perception and reasoning. As shown in Table \ref{tab:token_comparison}, this imposes a severe computational burden. For instance, on HRBench, VL-Rethinker increases token consumption by \textbf{157.2\%} compared to the base model, directly leading to high inference latency. Interestingly, while Monet is designed as a latent reasoning framework, it still incurs a substantial computational overhead. On BLINK, Monet uses \textbf{118.3 tokens} on average. Although this is a reduction compared to the base model, it remains orders of magnitude heavier than other latent approaches, suggesting that its ``visual thoughts'' still occupy a dense latent sequence.

In contrast, \textbf{Laser} achieves exceptional efficiency by shifting the reasoning process from the discrete token space to the continuous latent space, significantly outperforming both explicit and latent baselines. On the BLINK benchmark, Laser reduces the average token count to merely \textbf{6.0 tokens}, a reduction of \textbf{97.3\%}. This makes Laser substantially more efficient than Monet (118.3 tokens) and even surpasses the comparable latent method LVR (8.0 tokens). 

Crucially, as noted in Table~\ref{tab:results}, this efficiency does not compromise performance. While LVR suffers from semantic degradation due to its strict reconstruction objective, Laser's superposition mechanism allows it to encode richer, non-collapsed semantic information within the same compact latent budget. This confirms that Laser achieves a superior trade-off between efficiency and accuracy: it delivers the reasoning depth of explicit Chain-of-Thought models while maintaining the inference speed of direct-answer models. By condensing reasoning into a compact superposition state, Laser eliminates the need for generating hundreds of intermediate tokens, offering near-instantaneous inference suitable for real-time applications.

\noindent \textbf{RQ2: What is Laser's impact across variant tasks?}

\begin{figure}[htbp]
    \centering
    \includegraphics[width=0.7\linewidth]{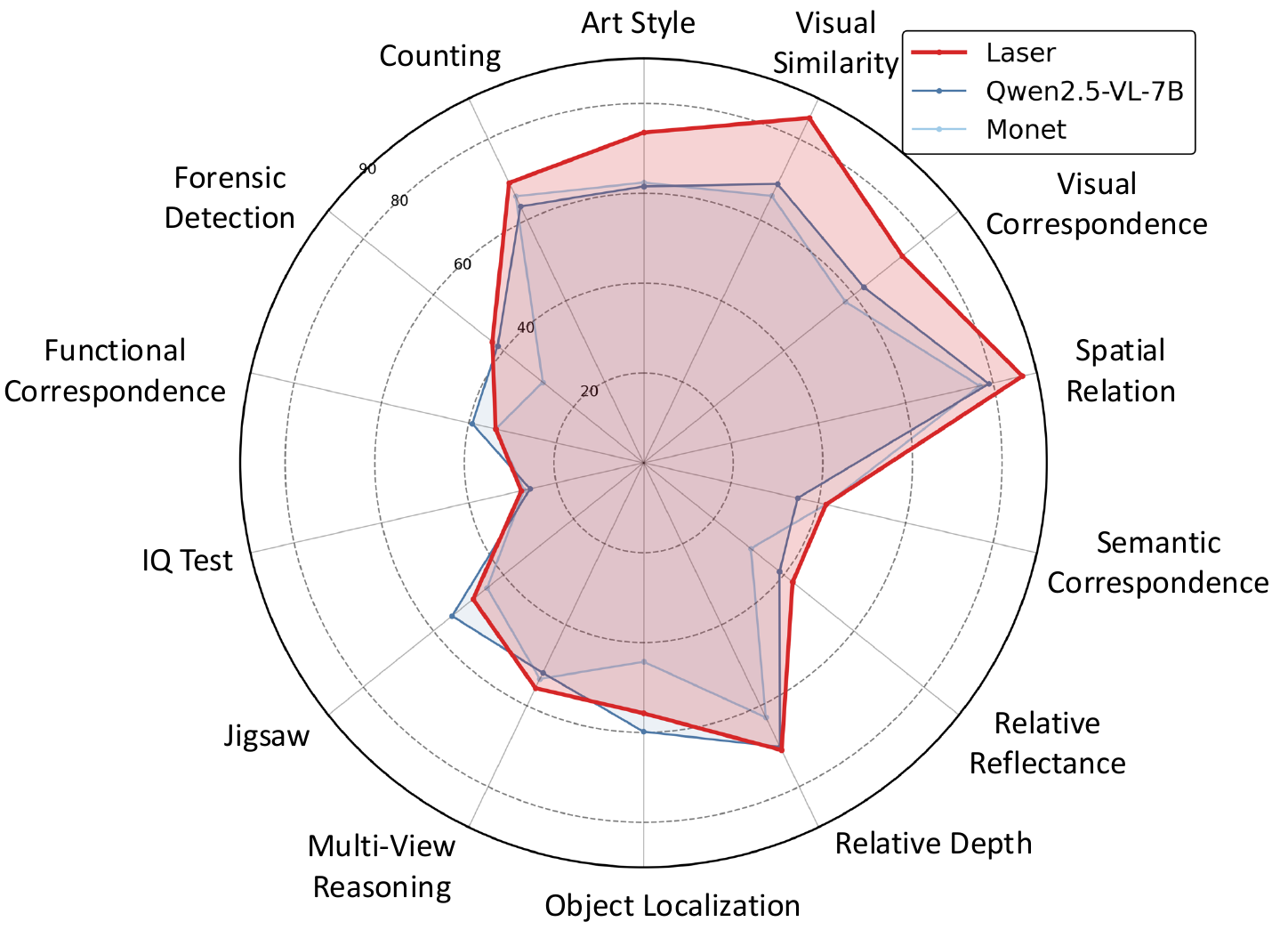}
    \caption{Fine-grained comparison across 14 distinct categories. Laser outperforms Qwen2.5-VL-7B and Monet in 11 tasks, highlighting superior high-level semantic and spatial reasoning.}
    \label{fig:rq2}
\end{figure}

We further conducted an in-depth experimental analysis of Laser across a range of task domains covering perception, understanding, and reasoning. As shown in Figure \ref{fig:rq2}, Laser exhibits a distinct performance profile: it achieves dominant superiority in 11 out of 14 tasks, with the only exceptions being \textit{Object Localization}, \textit{Jigsaw}, and \textit{Functional Correspondence}. This polarization reveals a fundamental insight into the "Forest Before Trees" cognitive mechanism.
Laser excels in tasks demanding high-level discrimination, such as \textit{Visual Similarity} and \textit{Spatial Relation}, by avoiding the premature information loss inherent in rigid tokenization. By preserving a "probabilistic haze" of visual nuances, the model effectively soft-matches ambiguous patterns and grasps relative 3D geometry. This confirms that our Dynamic Semantic Window successfully captures the holistic "Forest" structure, prioritizing scene-level contextual coherence over isolated object identification.
Conversely, Laser slightly underperforms in \textit{Object Localization} and \textit{Jigsaw}. This creates an intriguing contrast: while Laser is superior at understanding spatial relations, it is less precise at absolute pixel-level grounding. We attribute this to our weakly-supervised design choice. Unlike methods that rely on explicit bounding box regression (Region-of-Interest supervision), Laser learns grounding implicitly via latent alignment. It prioritizes the semantic flow of reasoning over rigid pixel reconstruction. Tasks like \textit{Jigsaw} require exact low-level feature matching rather than abstract semantic reasoning. Laser's ``Forest-first'' strategy naturally favors holistic scene understanding, leading the model to abstract away fine-grained pixel details. While this results in a minor trade-off in absolute localization precision, it yields the significant robustness observed in complex reasoning scenarios, mimicking human cognition, which is often semantically precise but metrically approximate.

\noindent \textbf{RQ3: Does Laser influence general vision-language abilities?}
\begin{table}[htbp]
    \centering
    \resizebox{\linewidth}{!}{
    \begin{tabular}{lcccccc}
        \toprule
        \textbf{Model} & \textbf{Multi-View Reasoning} & \textbf{Relative Depth} & \textbf{Geometry} & \textbf{Math} & \textbf{Web} & \textbf{Chart} \\
        \midrule
        
        Qwen2.5-VL-7B & 51.88 & 70.16 & 53.24 & 66.00 & 75.45 & 61.98 \\
        \textbf{Laser (Ours)} & \textbf{55.64} & \textbf{70.97} & 53.24 & \textbf{67.20} & \textbf{83.48} & \textbf{67.16} \\
        \bottomrule
    \end{tabular}
    } 
    
    \caption{We compare Laser against Qwen2.5-VL-7B across three dimensions: \textit{Spatial Perception} (Multi-View, Depth, Geometry), \textit{Visual Logic} (Math), and \textit{Structural Understanding} (Web, Charts). The results confirm that Laser successfully transfers reasoning patterns to unseen domains while preserving general spatial capabilities}
    \label{tab:generalization_results}
\end{table}

Addressing the pervasive risk of catastrophic forgetting, we evaluate whether Laser maintains its general-purpose visual foundation. The results on out-of-distribution domains\footnote{Task sources: Multi-View and Relative Depth (\textbf{Blink}), Geometry (\textbf{Geo}), Math (\textbf{MMStar}), and Web/Chart (\textbf{SEED-Bench-2+}).}confirm that our method avoids performance degradation and effectively preserves general skills. Specifically, Laser achieves substantial gains of 8.03\% on Web and 5.18\% on Chart tasks while maintaining a steady score of 53.24 on Geometry, suggesting that our "Forest-before-Trees" hierarchy reinforces the grasp of global structures. Furthermore, the learned deductive patterns successfully transfer to unseen logic domains, evidenced by improvements of 1.20\% in Math and 0.81\% in Relative Depth. As shown in Table~\ref{tab:generalization_results}, Laser boosts specialized reasoning without compromising the model's robust general capabilities.

\noindent \textbf{RQ4: Is Laser's reasoning interpretable?}

\begin{figure}[htbp]
    \centering
    \includegraphics[width=0.6\linewidth]{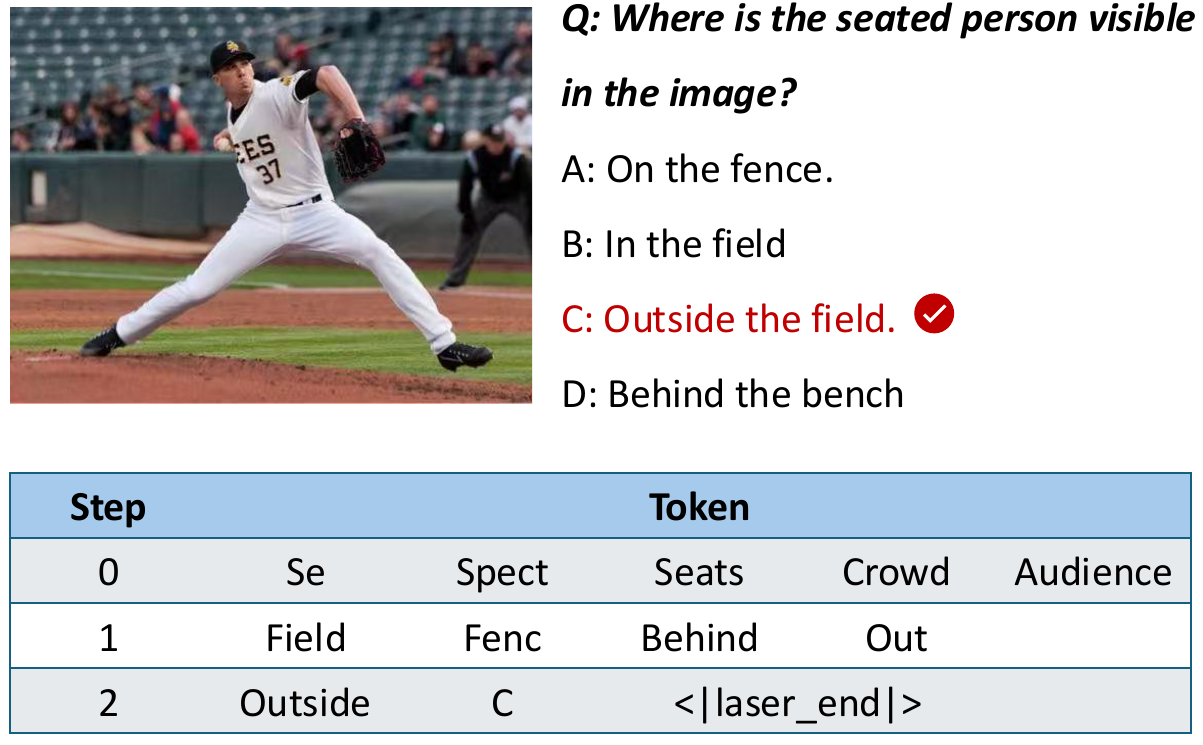}
    \caption{Visualization of the latent cognitive trajectory. The decoded tokens reveal a structured multi-hop reasoning path, evolving from entity localization (Step 0: Seats) to spatial analysis (Step 1: Fence) and final deduction.}
    \label{fig:r4}
\end{figure}

A pivotal advantage of Laser is its \textbf{inherent interpretability}, derived from the rigorous alignment between the visual projector and the LLM's semantic space. Unlike the opaque continuous vectors found in standard latent reasoning models, Laser's hidden states can be directly projected onto the vocabulary via the frozen LM head. This allows us to inspect the top-$k$ tokens at each intermediate step, effectively visualizing the model's ``cognitive trajectory.''
We demonstrate this capability using a representative case from MMStar, where the model must locate ``the seated person'' in a baseball scene dominated by a salient pitcher in the foreground.
\textbf{Initially (Step 0)}, the decoded tokens, led by ``Se-'' (suggesting \textit{Seats}), ``Spect-'' (\textit{Spectators}), and ``Crowd'', reveal that the model successfully overcomes visual saliency bias, shifting its attention to the background audience.
\textbf{Subsequently (Step 1)}, the latent state evolves to encode spatial constraints, with tokens such as ``Fence,'' ``Behind,'' and ``Out'' emerging to define the boundary between the spectators and the field.
\textbf{Finally (Step 2)}, this spatial reasoning converges into a semantic decision, as the probability mass shifts to ``Outside'' and the correct option label ``C.''
This trajectory confirms that Laser performs \textit{explicit-like} multi-hop reasoning, progressing from entity localization to spatial analysis and final deduction, entirely within the compact latent space.

\noindent \textbf{RQ5: How critical is each component?}
\begin{figure}[htbp]
    \centering
    \includegraphics[width=0.7\linewidth]{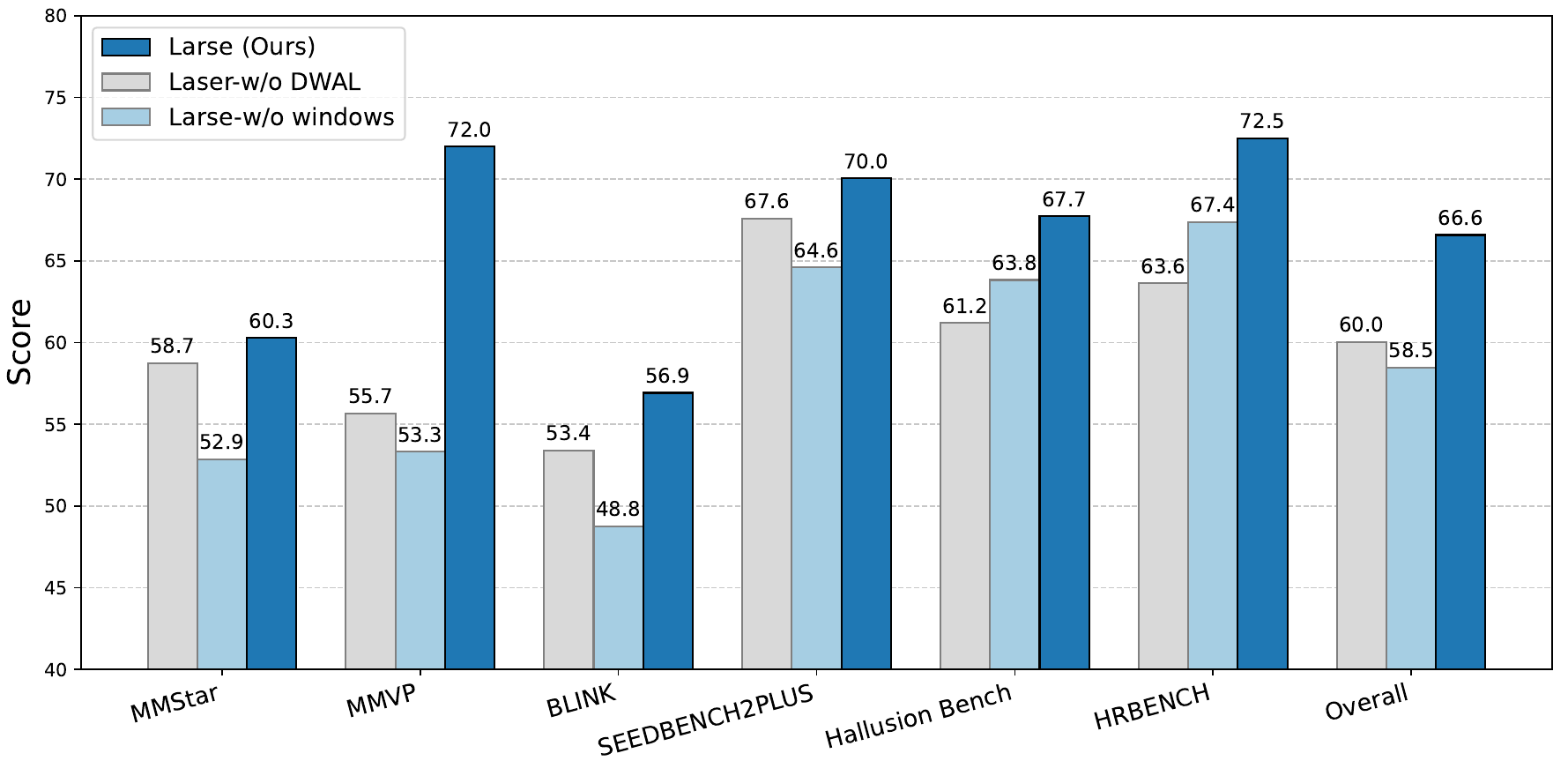}
    \caption{Ablation study. We contrast the full Laser model with variants lacking the DWAL (w/o DWAL) and the dynamic windows (w/o Windows). The consistent performance gap across six benchmarks highlights the necessity of the proposed Dynamic Windowed Alignment Learning for effective visual reasoning.}
    \label{fig:r5}
\end{figure}

To investigate the mechanisms behind the performance gains of Laser, we conduct an ablation study isolating the superposition objective and the windowing strategy. As illustrated in Figure~\ref{fig:r5}, the results reveal that these components serve distinct roles across different task types.
First, removing the DWAL objective (reverting to standard next-token prediction) leads to a significant drop in fine-grained perception benchmarks. This empirically validates that probabilistic superposition is crucial for preventing the ``premature semantic collapse'' often observed in standard autoregressive models.
Second, using a fixed validity window (``w/o Windows'') primarily impairs performance on complex reasoning tasks, with less impact on pure perception. This confirms that the Dynamic Window strategy is essential for enforcing the ``Forest-before-Trees'' hierarchy. By progressively shrinking the semantic scope, it ensures the model captures global context before focusing on local details.

\section{Conclusion}
We introduce \textbf{\textit{Laser}}, a method that transcends discrete Chain-of-Thought via continuous latent superposition. By reformulating visual deduction with Dynamic Windowed Alignment Learning, we enforce a "Forest-before-Trees" hierarchy that prevents premature semantic collapse. Laser achieves state-of-the-art performance among latent reasoning methods with superior robustness and a 97\% reduction in inference overhead. Notably, it stands as the first interpretable latent reasoning approach, suggesting that emancipating reasoning from rigid tokenization fosters native multimodal intelligence.

\bibliography{references}
\bibliographystyle{unsrt}

\newpage
\appendix

\addcontentsline{toc}{section}{Appendix}
\part{Appendix} 
\parttoc 

\section{Implementation Details}
\label{app:implementation}

The model is fine-tuned for 320 steps on 8 MI210 GPUs using 8 gradient accumulation steps. We utilize the AdamW optimizer with $\epsilon=1\mathrm{e}{-6}$ and a weight decay of $0.1$. The learning rate is initialized at $1\mathrm{e}{-5}$ with a cosine decay scheduler and a warmup ratio of $0.03$. To optimize memory efficiency, we employ DeepSpeed ZeRO-3 with CPU offloading and enable Flash Attention 2. We adopt a dynamic batching strategy with a token cap of 8,192 and a maximum per-device batch size of 16. For the proposed Laser objective, when the entropy intervention is triggered, the mixing coefficient for the hard target is set to $\alpha=0.8$. The input image resolution is dynamic, ranging from 128 to 8,192 tokens (approx. 100K to 6.4M pixels).

\section{Baseline Details}
\label{app:baseline}
\textbf{Vision-R1}~\cite{huang2025vision}. Vision-R1 explores R1-style post-training for MLLMs by constructing a large multimodal chain-of-thought cold start dataset and then applying RL with strategies such as progressive thinking suppression to improve multimodal reasoning.

\vspace{0.3em}
\noindent \textbf{PAPO}~\cite{wang2025perception}. PAPO (Perception-Aware Policy Optimization) targets the perception bottleneck in multimodal RLVR by introducing an implicit perception loss (KL term) that can be plugged into GRPO/DAPO, improving vision-dependent tasks without extra reward models or teacher models.

\vspace{0.3em}
\noindent   \textbf{DeepEyes}~\cite{zheng2025deepeyes}. DeepEyes studies interleaved multimodal reasoning (“thinking with images”) and incentivizes tool-assisted visual reasoning behaviors via end-to-end RL, using tailored data selection and reward design to improve grounding and reduce hallucinations.

\vspace{0.3em}
\noindent   \textbf{Monet}~\cite{wang2025monet}. Monet enables latent visual reasoning by generating continuous visual embeddings as intermediate “visual thoughts” and proposes a multi-stage distillation SFT pipeline plus visual-latent policy optimization to better train reasoning in latent visual space.

\vspace{0.3em}
\noindent   \textbf{VL-Rethinker}~\cite{wang2025vl}. VL-Rethinker enhances slow-thinking for VLMs via RL by adapting GRPO with selective sample replay and enforcing explicit self-reflection steps, improving multi-step multimodal reasoning performance.

\vspace{0.3em}
\noindent  \textbf{LLaVA-OneVision}~\cite{li2024llava}. LLaVA-OneVision is a unified open multimodal model designed to perform well across single-image, multi-image, and video scenarios, with strong cross-scenario transfer from images to videos.

\vspace{0.3em}
\noindent   \textbf{InternVL3.5-8B}~\cite{wang2025internvl3}. InternVL3.5 is a family of open-source multimodal models that improves reasoning and efficiency using cascade RL training and dynamic visual resolution routing; we adopt the 8B variant as a representative strong open baseline.

\vspace{0.3em}
\noindent   \textbf{Qwen2.5-VL-7B}~\cite{bai2025qwen2}. Qwen2.5-VL is a flagship open VLM series with improved recognition, localization, and document/video understanding; it introduces dynamic-resolution processing and absolute time encoding for long-video comprehension. We use the 7B model as the base open VLM baseline.

\vspace{0.3em}
\noindent  \textbf{LVR}~\cite{li2025latent}. Latent Visual Reasoning (LVR) enables autoregressive reasoning in the visual embedding space by training the model to generate latent states that reconstruct key visual tokens, interleaved with text generation; it can be further combined with RL to balance latent reasoning and textual outputs.

\section{Benchmark Details}
\label{app:benchmark}
\noindent \textbf{MMStar}~\cite{mmstar} is a vision-indispensable multimodal benchmark with carefully curated and purified samples to reduce language-only shortcuts and data contamination. It contains 1.5K questions and evaluates LVLMs over multiple core capabilities with fine-grained axes, providing a balanced diagnosis of both perception and reasoning.

\vspace{0.3em}
\noindent \textbf{MMVP}~\cite{mmvp} is constructed from 150 CLIP-blind image pairs and probes whether models can truly discriminate basic visual patterns that are obvious to humans but challenging for CLIP-like embeddings. It emphasizes failures where models produce high-confidence yet incorrect answers, often accompanied by plausible-sounding but ungrounded rationales.

\vspace{0.3em}
\noindent \textbf{BLINK}~\cite{blink} is a perception-centric benchmark with 3.8K multiple-choice questions, designed to test visual skills that humans can solve “within a blink,” such as correspondence, depth/geometry cues, forensics, and multi-view reasoning. Its tasks are intentionally difficult to solve from language priors alone, thereby stressing image-dependent perception.

\vspace{0.3em}
\noindent \textbf{SEEDBench2Plus}~\cite{seedbench2plus} targets text-rich visual comprehension in real-world formats (Charts, Maps, Web pages). It uses human-annotated multiple-choice questions to assess whether MLLMs can robustly read and reason over dense visual-text content.

\vspace{0.3em}
\noindent \textbf{Hallusionbench}~\cite{hallusionbench} is a diagnostic benchmark for entangled language hallucination and visual illusion in LVLMs. It uses human-crafted question structures with a control-group design to quantify logical consistency and common hallucination/illusion failure modes.

\vspace{0.3em}
\noindent \textbf{HRBench}~\cite{hrbench} evaluates fine-grained perception on high-resolution (4K/8K) images, where downsampling typically removes crucial details. It systematically tests HR visual understanding (including fine-grained single-/cross-instance perception) to measure whether models can handle true HR content.

\section{RL Analysis}
\label{sec:rl_analysis}

While our proposed method, Laser, demonstrates robust performance in the supervised fine-tuning stage, we further explore a reinforcement learning framework to align training behavior with inference dynamics and enable an autonomous ``early exit'' mechanism. To this end, we introduce the \textbf{EPG-GRPO} (Expected Policy Gradient - Group Relative Policy Optimization) algorithm. This framework integrates a variance-reduced gradient estimator with a length-invariant policy optimization scheme, successfully guiding the model to significantly reduce token consumption while maintaining performance parity.

\subsection{Optimization Objective}

Our objective combines Expected Policy Gradients (EPG) to stabilize learning within the high-variance latent space and a modified GRPO formulation to eliminate length bias.

Within LVR regions, standard single-token sampling often leads to gradient instability due to semantic ambiguity. To mitigate this, we calculate the gradient over the expectation of the Top-P subspace ($S_{\text{TopP}}$) rather than a single sampled token. We define the importance ratio $\rho_w$ for each token $w \in S_{\text{TopP}}$ and compute the expected surrogate loss $\mathcal{L}_{\text{epg}}^{(t)}$ as follows:
\begin{equation}
    \rho_w = \frac{\pi_{\theta}(w|s_t)}{\pi_{\text{old}}(w|s_t) + \epsilon},
\end{equation}

\begin{equation}
    \mathcal{L}_{\text{epg}}^{(t)} = \mathbb{E}_{w \in S_{\text{TopP}}} \left[ \min \left( \rho_w A_t, \text{clip}(\rho_w, 1-\epsilon, 1+\epsilon) A_t \right) \right],
\end{equation}
where $A_t$ is the advantage derived from group sampling.

For global optimization, standard GRPO normalizes by sequence length, which inadvertently incentivizes verbosity. We address this by standardizing the global loss using a fixed maximum completion length $L_{\text{max}}$, independent of the actual trajectory length $T_i$:
\begin{equation}
    \mathcal{L}_{\text{policy}} = \frac{1}{B \times L_{\text{max}}} \sum_{i=1}^{B} \sum_{t=1}^{T_i} \mathcal{L}_{\text{token}}^{(i,t)} + \beta \cdot \text{KL}(\pi_{\text{ref}} \| \pi_{\theta}).
\end{equation}
Here, $\mathcal{L}_{\text{token}}$ utilizes the derived $\mathcal{L}_{\text{epg}}$ for LVR steps and the standard clipped surrogate loss for explicit tokens.

\subsection{Exploration Strategy}

We employ a composite exploration strategy to prevent convergence to local optima and robustly learn termination conditions. First, we implement \textbf{Relative Norm Perturbation} to enhance the diversity of deterministic LVR hidden states. We inject Gaussian noise scaled by the signal's norm during the forward pass. For a hidden state $\mathbf{h}$, the perturbed state $\mathbf{h}'$ is given by:
\begin{equation}
    \mathbf{h}' = \mathbf{h} + \lambda \frac{\|\mathbf{h}\|}{\|\bm{\epsilon}\|} \bm{\epsilon}, \quad \bm{\epsilon} \sim \mathcal{N}(0, \mathbf{I}),
\end{equation}
where we set the noise ratio $\lambda = 0.05$. This ensures the perturbation is significant yet non-destructive across varying signal magnitudes. Second, to encourage efficiency, we apply \textbf{Stochastic Horizon Truncation}. For a subset of the generated samples, the maximum allowed steps $T_{\text{max}}$ are randomly sampled from a range $[T_{\text{min}}, T_{\text{upper}}]$. This forces the model to attempt convergence within limited horizons, thereby learning to optimize the efficiency of its reasoning path without relying on fixed-length priors.

\subsection{Composite Reward Engineering}

The reward function $R_{\text{total}}$ acts as a multi-objective optimization signal, aggregating components to balance accuracy, structural validity, efficiency, and diversity:
\begin{equation}
    R_{\text{total}} = R_{\text{acc}} + R_{\text{fmt}} + R_{\text{eff}} + R_{\text{div}}.
\end{equation}

\paragraph{Accuracy Reward ($R_{\text{acc}}$).}
Since our evaluation relies principally on multiple-choice questions or exact answer matching, we assign a binary reward based on the correctness of the final output:
\begin{equation}
    R_{\text{acc}} = 
    \begin{cases} 
        1 & \text{if answer is correct}, \\
        0 & \text{otherwise}.
    \end{cases}
\end{equation}

\paragraph{Format Reward ($R_{\text{fmt}}$).}
This component enforces the structural integrity of the dynamic protocol. A reward $r_{\text{fmt}}$ is granted solely if the sequence includes the requisite start token, successfully triggers the autonomous termination token \texttt{<|laser\_end|>}, and correctly encloses the final result in answer tags.

\paragraph{Efficiency Bonus ($R_{eff}$).}
To incentivize the model to voluntarily "exit early" when confident, we introduce a dynamic efficiency bonus. This reward is conditional on three strict constraints: the answer must be correct, the trajectory must not be forcibly truncated by the system limit ($T_{max}$), and the output must be free of format anomalies.
\begin{equation}
    R_{eff} = 
    \begin{cases} 
        \beta_{base} - \lambda_{step} \cdot T_{actual} & \text{if Correct } \land \ T_{actual} < T_{max}, \\
        0 & \text{otherwise}.
    \end{cases}
\end{equation}
Here, $\beta_{base}$ represents the maximum potential bonus, and $\lambda_{step}$ is a penalty coefficient that reduces the reward linearly with each reasoning step used. Under this formulation, a correct answer obtained via system truncation yields zero bonus, explicitly encouraging the model to learn autonomous termination logic.

\paragraph{Diversity Penalty ($R_{div}$).}
To prevent "state stagnation"—where adjacent reasoning steps exhibit excessive semantic redundancy—we apply a penalty based on the squared cosine similarity between consecutive hidden states $\mathbf{h}_t$ and $\mathbf{h}_{t-1}$.
\begin{equation}
    R_{div} = -\lambda_{div} \cdot \frac{1}{T-1} \sum_{t=1}^{T-1} \left( \text{sim}(\mathbf{h}_t, \mathbf{h}_{t-1}) \right)^2.
\end{equation}
We utilize the squared similarity to impose a non-linear penalty that aggressively punishes high-similarity states while remaining tolerant of the minor correlations necessary for coherent reasoning flow. The term is weighted by $\lambda_{div}$ and normalized by trajectory length to ensure consistent scaling across varying reasoning depths.

\subsection{Experimental Analysis}

To validate the efficacy of the EPG-GRPO framework, we compare the supervised baseline (Laser) with its RL-enhanced variant (Laser + EPG). As shown in Table~\ref{tab:rl_comparison}, the results demonstrate that our strategy successfully balances computational efficiency with reasoning capability.

The most significant impact is observed in inference efficiency. The RL-enhanced model reduces the average number of generated tokens by approximately 50\% across dynamic benchmarks such as BLINK and HRBench. This substantial decrease confirms that the model effectively internalized the autonomous early exit mechanism incentivized by the efficiency bonus $R_{eff}$ and stochastic horizon truncation, learning that concise reasoning paths are often sufficient.

Crucially, this efficiency gain is achieved without compromising general performance. The overall accuracy remains stable, with the method showing particular robustness on hallucinations (HallusionBench) and multi-modal reasoning (MMStar). While minor regressions were observed in select sensitivity-heavy tasks, the Subspace-EPG objective successfully preserved the semantic richness of the latent space, preventing the catastrophic forgetting or mode collapse often associated with RL fine-tuning.

\begin{table}[htbp]
\centering
\setlength{\tabcolsep}{4pt}
\resizebox{\linewidth}{!}{%
\begin{tabular}{l|cc|ccccccc}
\toprule
\multirow{2}{*}{\textbf{Model}} & \multicolumn{2}{c|}{\textbf{Avg. Tokens} $\downarrow$} & \multicolumn{7}{c}{\textbf{Accuracy (\%)}} \\
\cmidrule(lr){2-3} \cmidrule(lr){4-10}
 & BLINK & HRBench & MMStar & MMVP & BLINK & SEED & Hallusion & HRBench & Overall \\
\midrule

Laser (Main)
& 6.03 & 5.74
& 60.27 & 72.00 & \textbf{56.92} & 70.05 & 67.72 & \textbf{72.50} & 66.58 \\

\rowcolor{blue!5}
Laser + EPG
& \textbf{3.36} & \textbf{2.87}
& \textbf{60.87} & 72.00 & 55.76 & \textbf{70.79} & \textbf{68.98} & 72.00 & \textbf{66.73} \\

\bottomrule
\end{tabular}
}
\caption{Comparison between the main model (Laser) and the RL-enhanced model (Laser + EPG). \textbf{Avg. Tokens} denotes the average number of generated tokens on BLINK and HRBench.}
\label{tab:rl_comparison}
\setlength{\tabcolsep}{3pt} 
\renewcommand{\arraystretch}{1.1}
\end{table}

\section{Threshold Analysis for an Entropy-Adaptive Mechanism}

\begin{table}[htbp]
\centering
\definecolor{lightblue}{RGB}{223,234,242}
\setlength{\tabcolsep}{4pt}
\resizebox{\linewidth}{!}{
\begin{tabular}{lc cccccc c}
\toprule
\textbf{Threshold} ($\eta$) & \textbf{Trigger} & \textbf{MMStar} & \textbf{MMVP} & \textbf{BLINK} & \textbf{SEED} & \textbf{Hallusion} & \textbf{HRBench} & \textbf{Overall} \\
\midrule

$\eta=1.0$ & 0.0\% 
& 59.93 & 69.00 & 56.86 & 69.52 & 67.93 & 71.88 & 65.85 \\

$\eta=0.8$ & 2.5\% 
& \textbf{60.33} & 69.67 & 56.08 & 69.96 & \textbf{68.24} & 71.75 & 66.01 \\

\rowcolor{lightblue}
$\eta=0.6$ (Ours) & 10.0\% 
& 60.27 & \textbf{72.00} & \textbf{56.92} & \textbf{70.05} & 67.72 & 72.50 & \textbf{66.58} \\

$\eta=0.5$ & 18.0\% 
& 57.40 & 71.67 & 54.97 & 68.95 & 64.56 & \textbf{72.75} & 65.05 \\

\bottomrule
\end{tabular}
} 

\caption{Ablation study on the entropy threshold $\eta$. ``Trigger'' denotes the intervention activation ratio. $\eta=0.6$ yields the best balance and is used as our default setting.}
\label{tab:ablation_eta}
\end{table}

\subsection{Analysis of Entropy Threshold ($\eta$)}

We first analyze the impact of the entropy threshold $\eta$ by correlating it with the \textbf{Trigger Ratio}, defined as the percentage of tokens where the model's high uncertainty necessitates a hard teacher intervention. As shown in Table~\ref{tab:ablation_eta}, optimal performance is achieved at $\eta=0.6$, corresponding to a trigger ratio of approximately 10\%. This suggests that intervening on roughly one in ten tokens provides sufficient grounding signals to correct the reasoning trajectory without disrupting the semantic flow. 

When the threshold is lowered to $\eta=0.5$, the trigger ratio rises to 18\%, imposing a stricter constraint akin to standard supervision. While this rigid guidance benefits tasks requiring precise alignment (yielding the highest score on HRBench), it stifles the latent exploration necessary for complex logic, resulting in performance degradation on reasoning-heavy benchmarks like MMStar (57.40) and HallusionBench (64.56). Conversely, higher thresholds ($\eta=0.8, 1.0$) lead to negligible intervention ($<2.5\%$). Although this preserves flexibility, it lacks the necessary corrective mechanism to fix visual grounding errors, leading to suboptimal results in fine-grained perception tasks such as MMVP. Thus, $\eta=0.6$ strikes an effective balance, enforcing visual validity while maintaining cognitive flexibility.

\begin{table}[htbp]
\centering
\definecolor{lightblue}{RGB}{223,234,242}
\setlength{\tabcolsep}{4pt} 

\resizebox{0.9\linewidth}{!}{
\begin{tabular}{l cccccc c}
\toprule
\textbf{Threshold} ($\alpha$) & \textbf{MMStar} & \textbf{MMVP} & \textbf{BLINK} & \textbf{SEED} & \textbf{Hallusion} & \textbf{HRBench} & \textbf{Overall} \\
\midrule

$\alpha=0.2$
& 60.00 & 70.00 & 55.50 & 69.65 & \textbf{68.66} & \textbf{72.62} & 66.07 \\

$\alpha=0.5$
& 59.87 & \textbf{73.00} & 55.65 & 69.83 & 68.24 & 72.38 & 66.50 \\

\rowcolor{lightblue}
$\alpha=0.8$ (Ours)
& \textbf{60.27} & 72.00 & \textbf{56.92} & \textbf{70.05} & 67.72 & 72.50 & \textbf{66.58} \\

\bottomrule
\end{tabular}
}
\caption{Ablation study on the parameter $\alpha$. $\alpha \in [0, 1]$ controls the intensity of the hard intervention. We observe that $\alpha=0.8$ achieves the optimal balance.}
\label{tab:ablation_alpha}
\end{table}

\subsection{Analysis of Intervention Intensity ($\alpha$)}

We further examine the impact of the parameter $\alpha$, which modulates the intensity of the hard intervention once high uncertainty is detected. As indicated in Table~\ref{tab:ablation_alpha}, increasing $\alpha$ leads to a consistent, albeit marginal, improvement in overall performance. We observe that a lower $\alpha$ (e.g., $0.2$) applies a softer correction, which appears insufficient to fully resolve ambiguity when the model is highly uncertain. Conversely, a higher setting ($\alpha=0.8$) provides a more decisive guidance signal, effectively acting as a ``hard reset'' to realign the latent trajectory with the ground truth. This suggests that during critical states of high uncertainty, prioritizing deterministic constraints over soft superposition yields slightly more robust reasoning.

\section{Time-Aware Semantic Decay}

We further explore a \textbf{Time-Aware Semantic Decay} component to regulate the semantic distribution within the validity window. This strategy applies a temporal bias to the target logits based on their relative distance from the current step:
\begin{equation}
    \tilde{z}_t^{(k)} = \hat{z}_t^{(k)} + \ln (\gamma^{k-t}),
\end{equation}
where $k \in W_t$ denotes the token index and $\gamma \in (0, 1]$ acts as a decay factor. This formulation allows the framework to flexibly modulate the attention density assigned to distant future tokens relative to immediate deductive steps.
\label{sec:time_decay}
\begin{table}[htbp]
\centering
\definecolor{lightblue}{RGB}{223,234,242}
\setlength{\tabcolsep}{6pt}

\resizebox{\linewidth}{!}{
\begin{tabular}{l cccccc c}
\toprule
\textbf{Model} & \textbf{MMStar} & \textbf{MMVP} & \textbf{BLINK} & \textbf{SEED} & \textbf{Hallusion} & \textbf{HRBench} & \textbf{Overall} \\
\midrule

\rowcolor{lightblue}
Larse (Default)
& \textbf{60.27} & 72.00 & \textbf{56.92} & 70.05 & \textbf{67.72} & \textbf{72.50} & \textbf{66.58} \\

w/ Time Decay
& 59.60 & \textbf{73.00} & 55.34 & \textbf{70.18} & 67.40 & 72.38 & 66.32 \\

\bottomrule
\end{tabular}
}
\caption{Ablation study comparing the standard Laser model with a time-aware variant.}
\label{tab:ablation_time_decay}
\end{table}
\section{Prompt Engineering}
\label{app:prompt}

To ensure the synthesis of high-quality cognitive scanpaths, we employed the following structured system prompt. It enforces a strict ``Global-to-Local'' scanning logic and output format constraints.

\begin{tcolorbox}[
    colback=gray!5,      
    colframe=gray!60,    
    boxrule=0.5pt,       
    arc=2mm,             
    left=2pt, right=2pt, top=2pt, bottom=2pt, 
    fontupper=\scriptsize\ttfamily\raggedright 
]

\textbf{\#\#\# Role Definition}\\
You are a \textbf{Visual Cognitive Engine} designed to deconstruct the visual reasoning process into a strictly ordered \textbf{"Visual Scanpath"}.

\vspace{0.5em}
\textbf{\#\#\# Core Objective}\\
Generate a sequential stream of \textbf{Atomic Visual Concepts}. This stream must represent a logical flow of discovery: scanning from the global environment, zooming into specific objects, and accumulating visual cues, culminating in the \textbf{most critical information} needed to answer the user's query.

\vspace{0.5em}
\textbf{\#\#\# Precision \& Format Principles (CRITICAL)}\\
1. \textbf{Atomic Specificity:} Be as specific as the image clarity allows \textbf{immediately} (e.g., "Ferrari" not "Car"), but strictly use single words or 1-3 word phrases.\\
2. \textbf{De-Grammatized Output:} Output dense information only. \textbf{REMOVE} all stop words (is, the, a, of, in).\\
3. \textbf{Visual Certainty:} Only output concepts that are \textbf{visually observable}. Use broader terms if blurry.\\
4. \textbf{Contextual Anchoring:} Repeat a previous entity ONLY if necessary to attach the final resolution.

\vspace{0.5em}
\textbf{\#\#\# The 4-Stage Scanning Logic (Strict Order)}\\
\textbf{1. Global Anchor (Step 1-2):} Start with the broadest visible context (e.g., "Kitchen", "Blue Sky").\\
\textbf{2. Subject Localization (Step 3-X):} Locate the main subject relevant to the question.\\
\textbf{3. Visual Evidence (Step X-Y):} List visible attributes or actions supporting the answer.\\
\textbf{4. Critical Resolution (Final Step):} The specific concept answering the query must appear at the very end.

\vspace{0.5em}
\textbf{\#\#\# Negative Constraints}\\
- \textbf{NO Premature Reveals:} Do not output the answer early.\\
- \textbf{NO Artificial Hierarchy:} No "Fruit" $\to$ "Apple", just "Apple".\\
- \textbf{NO Sentences:} Raw concepts only.

\vspace{0.5em}
\textbf{\#\#\# Output Format}\\
Output strictly valid JSON:\\
\{\\
\ \ "reasoning\_chain": [\\
\ \ \ \ "String1",\\
\ \ \ \ "String2",\\
\ \ \ \ ...\\
\ \ ]\\
\}

\end{tcolorbox}

\section{Dataset Details}
\label{app:data_details}

\begin{table}[htbp]
    \centering
    \resizebox{0.8\linewidth}{!}{
    
    \begin{tabular}{llccccc}
        \toprule
        \multirow{2}{*}{\textbf{Source}} & \multirow{2}{*}{\textbf{Task Domain}} & \multicolumn{2}{c}{\textbf{Data Scale}} & \multicolumn{3}{c}{\textbf{Reasoning Nodes}} \\
        \cmidrule(lr){3-4} \cmidrule(lr){5-7}
         & & Count & Ratio & Min & Max & Mean \\
        \midrule
        
        \textbf{Flickr30k} & Captioning & 103,790 & 38.5\% & 2 & 20 & 7.50 \\
        \textbf{GQA} & VQA/Reasoning & 86,218 & 32.0\% & 2 & 18 & 7.13 \\
        \textbf{OpenImages} & Detection & 42,639 & 15.8\% & 3 & 17 & 6.38 \\
        \textbf{Visual7W} & VQA & 30,271 & 11.2\% & 1 & 17 & 6.85 \\
        \textbf{CUB} & Fine-grained Cls. & 3,521 & 1.3\% & 4 & 18 & \textbf{8.55} \\
        \textbf{VSR} & Spatial Reasoning & 3,334 & 1.2\% & 2 & 16 & 7.08 \\
        \midrule
        \textbf{Total / Avg.} & - & \textbf{269,773} & \textbf{100\%} & \textbf{1} & \textbf{20} & \textbf{7.14} \\
        \bottomrule
    \end{tabular}
    } 
    
    \caption{Detailed statistics of the \textbf{\textit{ScanPath}} dataset. The dataset integrates six diverse visual tasks. \textbf{Nodes} refers to the number of discrete semantic anchors in the reasoning chain (distinct from token length).}
    \label{tab:data_stats}
\end{table}

Table~\ref{tab:data_stats} outlines the detailed statistics of the \textbf{\textit{ScanPath}} dataset, comprising 269,773 samples across six visual domains. We measure chain complexity using ``Reasoning Nodes''—defined as discrete semantic anchors (e.g., region identification or attribute verification) rather than linguistic token length. As shown in the table, the node distribution naturally aligns with task difficulty: fine-grained tasks like CUB-200 require deeper reasoning paths (Mean: 8.55 nodes) compared to basic detection tasks like OpenImages (Mean: 6.38 nodes), resulting in a balanced overall average of 7.14 nodes.

\section{Details of Human Annotations}

To assess the quality of our dataset, we recruited three expert annotators (Ph.D. candidates in Computer Science) to manually evaluate 200 randomly sampled instances from \textit{ScanPath}. The evaluation focused on two key dimensions: (1) the validity of the visual reasoning chain, and (2) adherence to the global-to-local logic. The results demonstrated a validity rate of 91.5\% with substantial inter-annotator agreement (Fleiss' $\kappa = 0.677$), confirming the reliability of our automated pipeline and filtering protocol.

\section{Case Study}

We present test cases of our method during inference, as illustrated in Figure xx. These cases clearly demonstrate the efficiency and accuracy of our method during the reasoning process.

\begin{figure*}[htbp]
    \centering
    \includegraphics[width=\linewidth]{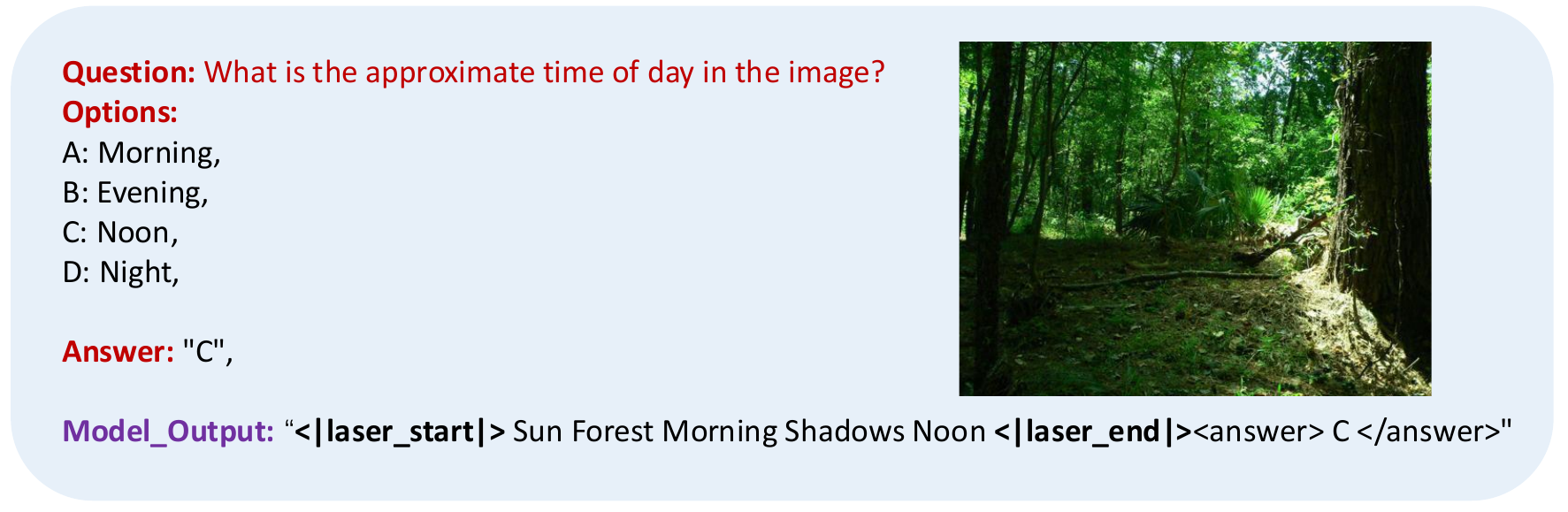}
    \caption{A test case from MMStar showcases the efficacy and efficiency of our Laser.}
    \label{fig:case1}
\end{figure*}

\begin{figure*}[htbp]
    \centering
    \includegraphics[width=\linewidth]{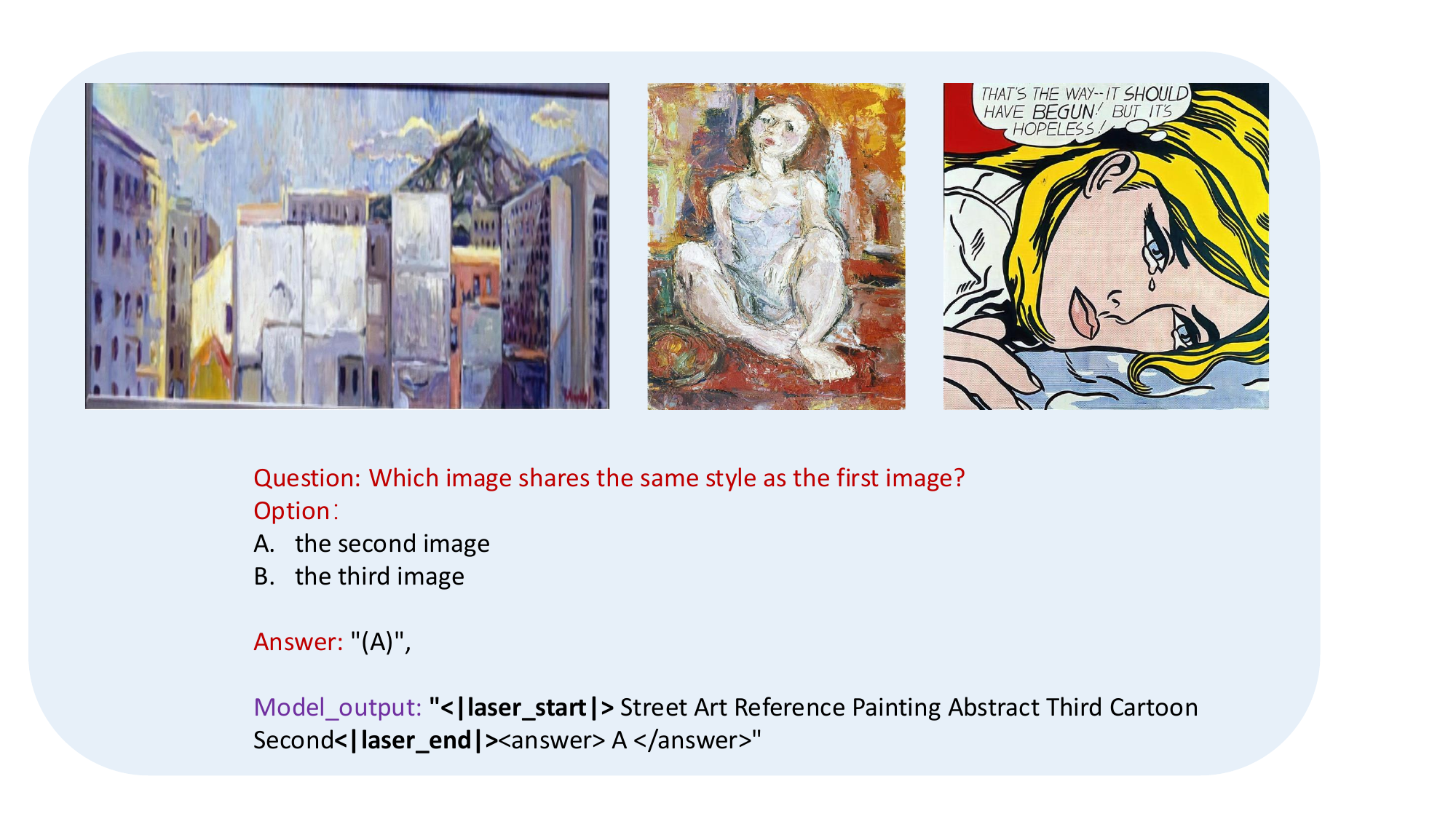}
    \caption{This multi-image reasoning test case from MMStar illustrates the effectiveness and efficiency of our Laser.}
    \label{fig:rq2}
\end{figure*}


\newpage

\end{document}